\documentclass{ecai} 

\usepackage{url}
\usepackage{latexsym}
\usepackage{amssymb}
\usepackage{amsmath}
\usepackage{amsthm}
\usepackage{booktabs}
\usepackage{enumitem}
\usepackage{graphicx}
\usepackage{color}
\usepackage{multirow}

\newcommand{\BibTeX}{B\kern-.05em{\sc i\kern-.025em b}\kern-.08em\TeX}

\begin{document}


\begin{frontmatter}


\paperid{123} 


\title{Informed Learning for Estimating Drought Stress at Fine-Scale Resolution Enables Accurate Yield Prediction}


\author[A, B, C]{\fnms{Miro}~\snm{Miranda}
\thanks{Corresponding Author. Email: miro.miranda\_lorenz@dfki.de}}
\author[A]{\fnms{Marcela}~\snm{Charfuelan}
}
\author[C]{\fnms{Matias}~\snm{Valdenegro Toro}
} 
\author[A, B]{\fnms{Andreas}~\snm{Dengel}
} 
\address[A]{German Research Center for Artificial Intelligence, Kaiserslautern, Germany}
\address[B]{Department of Computer Science, University of Kaiserslautern-Landau, Kaiserslautern, Germany}
\address[C]{Bernoulli Institute, University of Groningen, Groningen, Netherlands}


\begin{abstract}
Water is essential for agricultural productivity. Assessing water shortages and reduced yield potential is a critical factor in decision-making for ensuring agricultural productivity and food security.
Crop simulation models, which align with physical processes, offer intrinsic explainability but often perform poorly. Conversely, machine learning models for crop yield modeling are powerful and scalable, yet they commonly operate as black boxes and lack adherence to the physical principles of crop growth.
This study bridges this gap by coupling the advantages of both worlds. We postulate that the crop yield is inherently defined by the water availability. Therefore, we formulate crop yield as a function of temporal water scarcity and predict both the crop drought stress and the sensitivity to water scarcity at fine-scale resolution. 
Sequentially modeling the crop yield response to water enables accurate yield prediction. To enforce physical consistency, a novel physics-informed loss function is proposed. 
We leverage multispectral satellite imagery, meteorological data, and fine-scale yield data. Further, to account for the uncertainty within the model, we build upon a deep ensemble approach. Our method surpasses state-of-the-art models like LSTM and Transformers in crop yield prediction with a coefficient of determination ($R^2$-score) of up to 0.82 while offering high explainability. 
This method offers decision support for industry, policymakers, and farmers in building a more resilient agriculture in times of changing climate conditions. 
The code is publicly available at \url{https://github.com/mmiranda-l/Yield-Loss}.
\end{abstract}

\end{frontmatter}


\section{Introduction}
\begin{figure*}[!t]
  \centering
  \includegraphics[width=.99\linewidth]{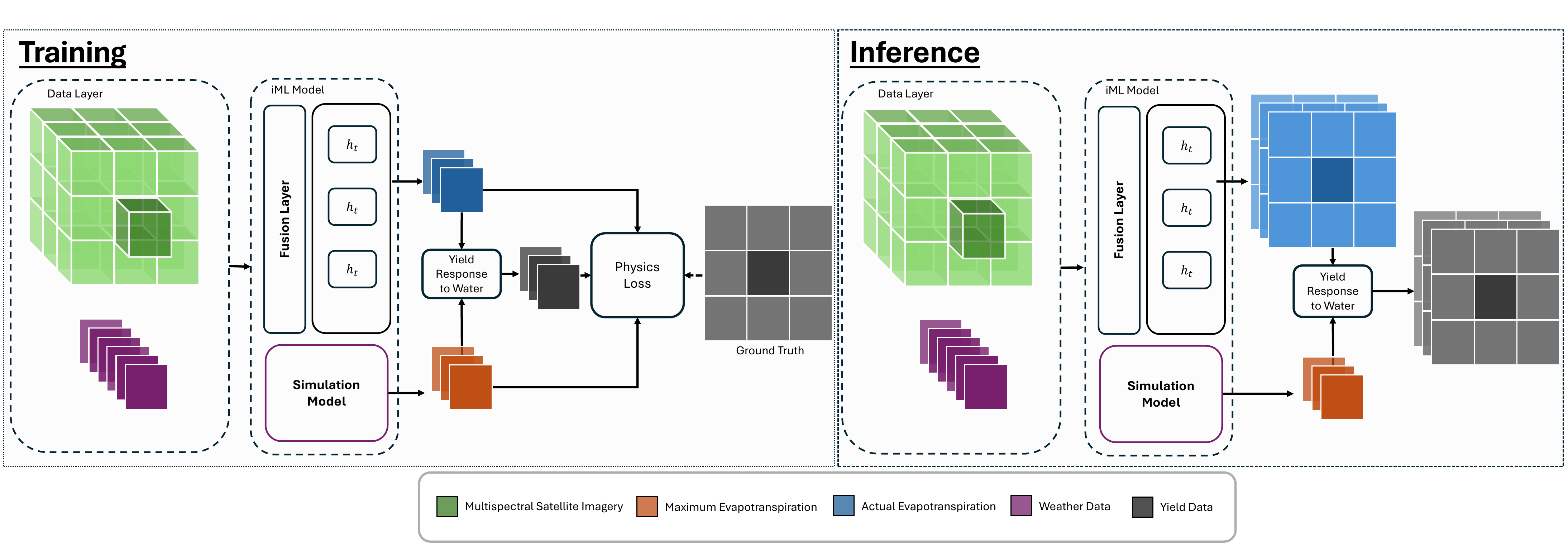}
  \caption{Overview of the physics-guided method for crop yield prediction. The training (\texttt{left}) and the inference (\texttt{right}) are displayed. \textbf{Highlight:} The Data is modeled pixel-wise by optimizing the model to approximate crop yield using the yield response to water function, which enhances the estimation of actual evapotranspiration, an indicator of drought stress. Both components are estimated at fine-scale resolution. In contrast, the simulated maximum evapotranspiration lacks spatial detail. }
  \label{fig:model_training_inference}
\end{figure*}%
Closing the gap between potential and actual yields is an urgent task to sustain global food security~\cite{fischer2012crop}. Extreme weather conditions like droughts and floodings are pressing challenges in the agricultural sector, directly affecting productivity and causing substantial yield and economic losses every year \cite{arora2019impact,molotoks2021impacts,malhi2021impact}. 
The response of crop yields to water scarcity has been a central focus of research for decades, serving as a critical parameter in assessing crop resilience under extreme weather conditions~\cite{pereira2015crop}. 
Traditionally, simulation models, also known as process-based models, have been commonly employed to capture this relationship. Simulation models build upon biological and physical principles and offer high explainability, supporting decision-making in areas such as irrigation, fertilization, and disease control. However, crop simulation models often struggle with large and multidimensional data, are computationally expensive, and require careful calibration. Therefore, the application of simulation models to large areas and high spatial resolution is limited. Furthermore, simulation models are typically simplified representations of reality, relying on approximations or reference environments to maintain computability~\cite{kang2009climate}. This can result in inaccurate performances \cite{leng2020predicting}. 
Therefore, to mitigate these limitations, machine learning (ML) models are increasingly utilized for crop productivity estimation~\cite{van2020crop}. Recent studies demonstrate impressive scalability and accuracy, even at fine-scale resolution~\cite{helber2023crop,pathak}. ML models handle complex and multidimensional data efficiently~\cite{perich2023pixel,mena2025adaptive}. However, ML models are commonly criticized for their black-box characteristics, limiting their transparency and explainability \cite{castelvecchi2016can,roscher2020explainable}. 
Additionally, ML models are seldomly designed to follow the underlying physical principles of plant growth~\cite{he2023physics}, which can cause a significant lack of trust and even invalid outcomes. Incorporating physical consistency remains essential for opening the black-box of ML models, building trust in ML-based predictions, and finally for decision-support in the agricultural sector. Consequently, there is a growing demand to merge the strengths of data-driven approaches with the interpretability of simulation models \cite{dash2022review,kang2009climate,roscher2020explainable,von2021informed}.\\

This study overcomes existing limitations by coupling interpretable simulation models and high-performance ML models. We formulate crop yield as a function of water scarcity and sequentially learn the actual \textit{evapotranspiration}, a proxy for drought and crop stress, and the crop susceptibility to water scarcity. This is used to derive the expected yield loss, by sequentially solving the crop yield response to water function \cite{doorenbos1979yield} at fine-scale resolution. Additionally, we enforce physical consistency using a novel physics-informed loss function. The crop drought stress is approximated at $10 \times 10 \text{ m}$ spatial resolution using multispectral satellite imagery from the Sentinel-2 mission and coarse weather data. 
In detail, we present the following contributions:
\begin{itemize} 
    \item We explicitly consider that crop yield is a function of water scarcity by incorporating the crop yield response to water function into the loss term. Thus, we outperform several state-of-the-art methods in crop yield prediction while demonstrating physical consistency and explainability.
    \item We demonstrate that the crop drought stress can be approximated at $10 \times 10 \text{ m}$ spatial resolution. This represents a significant improvement over existing simulation models, which commonly lack spatial detail. 
    \item We present a novel architecture based on a Long-Short-Term-Memory backbone with a temporal attention mechanism. Furthermore, to explicitly account for the uncertainty in the model, we leverage a deep ensemble approach. 
\end{itemize}
The results are demonstrated on a publicly available yield dataset for cereal crops, collected in Switzerland between 2017-2021. 


\section{Related Work}
Crop yield prediction using ML is an intensely studied field of research \cite{van2020crop}, especially in Earth Observation (EO). With studies focusing on various crop types, countries, model architectures, and input features. For instance,~\citet{helber2023crop} proposed an operational approach for crop yield prediction that is globally scalable for various crop types by relying solely on high-resolution multispectral satellite imagery and ground truth yield data with fine-scale resolution. This approach was subsequently extended by integrating additional data modalities, including weather, soil, and terrain elevation, under various data fusion schemes and time series representations~\cite{pathak,mena2025adaptive,miranda2024multi,miranda2025efficient}. 
Although, data-driven approaches represent a step into scalable crop yield prediction, they still operate as black-boxes and lack adherence to crop growth physical principles~\cite{he2023physics}. Therefore, \citet{najjar2025explainability} raised concerns about the transparency and interpretability of ML models in EO applications.\\
Several studies have focused on integrating prior knowledge into the learning algorithm, aiming to improve the model robustness, and explainability~\cite{roscher2020explainable}.~\citet{von2021informed}, provide a taxonomy for the explicit integration of prior knowledge into the learning pipeline, leading to the concept of informed machine learning (iML). 
The scientific consistency of a model's predictions is a fundamental requirement in the natural sciences, enforcing a plausible solution space that follows governing physical principles. Scientific consistency can be commonly enforced into ML models through regularization \cite{roscher2020explainable,von2021informed}, with physics-informed (PI) learning \cite{rudy2017data,raissi2019physics} being a prominent example. \citet{karpatne2017theory} argued that scientific consistency must be considered as a performance measure in ML when deploying models into practice, especially in safety-critical applications. \\
Studies exist that integrate prior knowledge to enhance crop yield prediction. Several of which acknowledge the importance of drought stress.
However, the majority of studies focus on enriching the data space with domain knowledge. 
For instance, \citet{shuai2022subfield} demonstrated a strong improvement by integrating a crop drought index into the data space of the ML model, thereby increasing the robustness to extreme weather conditions for maize yield prediction. Other studies that demonstrate improved performance by enriching the data space under extreme weather conditions are evidenced in \cite{shahhosseini2021coupling}. Likewise, \citet{jahromi2023developing} demonstrated that including the evapotranspiration into the data space is particularly important for improving crop yield predictions using ML. \\
Although several studies demonstrate the importance of domain knowledge, data space enrichment or data augmentation does not guarantee scientific consistency. 
In contrast, only a few studies exist that particularly enforce scientific consistency through regularization. For instance, \citet{he2023physics} presented a physics-guided approach for crop yield prediction by acknowledging key components of the carbon cycle while enforcing spatial fairness. Similarly, \citet{he2024knowledge} presents an approach to extract physics-aware features from simulation data to estimate crop yield while preserving the physical features. \\
To the best of our knowledge, no study exists that estimates the crop yield response to water scarcity at the pixel level from coarse and inaccurate simulation data while enforcing physical consistency and model interpretability. 
\section{Background}
Water shortages significantly impact crop development. Consequently, substantial effort is made to monitor and model water requirements to enable timely interventions in the event of a water shortage. In an agricultural context, this usually involves estimating the crop \textit{evapotranspiration (ET)}. 
\subsection{Evapotranspiration (ET)}
The \textit{evapotranspiration (ET)} is the sum of all biophysical processes in which liquid water is converted to water vapor from various surfaces, including topsoil (evaporation) and vegetation (transpiration). The ET is a direct indicator of crop health and crop stress \cite{allen2007satellite,jahromi2023developing} and is closely correlated with crop growth and crop yield \cite{allen2011evapotranspiration,khan2019estimating}. Consequently, ET is a core component for crop management, such as irrigation planning and crop water deficit mitigation, and is particularly important in water-scarce regions. 
The ET is reflected by various components, including temperature, solar radiation, soil, terrain elevation, and crop properties. For a reference environment, a detailed description of the biophysical processes is provided by the Food and Agriculture Organization (FAO)-56 method~\cite{allen1998crop}, using the Penman-Monteith equation, recommended for daily ET [mm d$^{-1}$] estimation: \\
\begin{equation}\label{eq:ET_o}
    ET_o = \left( \frac{0.408 \Delta (R_n - G) + \gamma \frac{900}{T + 273} u_2 (e_s - e_a)}{\Delta + \gamma \left(1 + 0.34 u_2\right) } \right),
\end{equation}
where:
\begin{description}
    \item $ET_o$: reference evapotranspiration [mm $\text{day} ^{-1}$]
    \item $R_n$: net radiation at the crop surface [MJ $\text{m}^{-2} \text{day}^{-1}$]
    \item $G$: soil heat flux density [MJ $\text{m}^{-2} \text{day}^{-1}$]
    \item $T$: mean daily temperature at 2m [°C]
    \item $u_2$: wind speed at 2m [m $\text{s}^{-1}$]
    \item $e_s$ and $e_a$: saturated and actual vapor pressure [kPa]
    \item $\Delta$ slope vapor pressure curve [kPa °C$^{-1}$]
    \item $\gamma$: psychrometric constant [kPa °C$^{-1}$]
\end{description}

In an agricultural context and for a specific crop type, \citet{allen1998crop} differentiates between the maximum ET ($ET_x$) and the actual ET ($ET_a$). 
The maximum ET is defined under standard and non-limiting environmental conditions and is solely defined by climate conditions and crop-specific parameters, achieving full productivity, such as disease-free, well-fertilized, and under optimum soil water conditions. The maximum ET is defined by:
\begin{equation}\label{eq:etx}
    ET_x = K_c \cdot ET_o, 
\end{equation}
where $K_c$ is a dimensionless coefficient, varying predominantly with crop-specific characteristics that distinguishes a specific crop type from the reference environment. Often $K_c$ is separated into a dual crop coefficient, one for crop transpiration and one for soil evaporation. In this study the dual crop coefficient is used. Note that the actual ET can be greater than the reference ET, depending on the $K_c$ values. \\
In contrast, the $ET_a$ represents actual ET under limiting environmental conditions caused by low water potential, resulting in water stress and a reduction in ET.
The actual ET is defined by:
\begin{equation}
    ET_a = K_s \cdot K_c \cdot ET_o,
\end{equation}
with $K_s$ being a water stress coefficient that is determined by the crop type and the growth stage. Usually, the stress coefficient only impacts the crop transpiration. 
This results in a reduction of the ET, and ultimately causing a reduction in crop productivity. For instance, $K_s < 1$ for water-limiting environments, and $K_s = 1$ otherwise. 
Various factors cause productivity-limiting conditions, including soil infertility, soil salinity, limited soil water content, diseases, and poor management. Nevertheless, the stress coefficient $K_s$ is commonly idealized by simulation models and therefore may not accurately reflect the actual conditions in the field, particularly in water-scarce regions \cite{allen2007satellite}.
However, in the light of extreme weather conditions, the frequency of severe droughts and floodings is expected to increase, causing either water scarcity or water abundance \cite{kang2009climate,arora2019impact}. Therefore, accurate modeling of crop water stress is a fundamental challenge in EO. 
\subsection{Evapotranspiration \& Yield Loss}
In an earlier work, the FAO described the relationship between ET and the relative yield loss~\cite{doorenbos1979yield}, stating that the relative reduction in ET is related to the relative reduction in yield:
\begin{equation}\label{eq:yieldl}
\begin{aligned}
y_l = \left(  1-\frac{y_a}{Y_x}\right) = K_y \left( 1- \frac{ET_a}{ET_x}\right)
\end{aligned}
\end{equation}
where:
\begin{description}
    \item $y_l$: relative yield loss [\%]
    \item $y_a$: actual yield
    \item $Y_x$: potential maximum yield 
    \item $K_y$: yield response factor 
\end{description}
The dimensionless yield response factor $K_y$ represents the effect of a reduction in ET on the crop yield by capturing the complex relationship between ET and productivity. More specifically, $K_y >1$ indicates high sensitivity to water deficits with a proportionally larger yield reduction, and $K_y < 1$ indicates higher resilience to water deficits. Different studies exist that have empirically estimated $K_y$ coefficients for various crops. However, often reporting discrepancies, making the equation difficult to solve in practice. Furthermore, $K_y$ values change over the growing period since many crops exhibit variable susceptibility to water scarcity over the growing period. This subsequently increases the difficulty of accurately estimating the yield response factor. An approximation for various crop types is given in~\cite{steduto2012crop}. For example, cereal crops are reported with a $K_y$ value of approximately 0.5 during vegetative phases and 1.5 during flowering with an average value of 1.05.

\subsubsection{Problem Definition \& Hypothesis}
More importantly, while $ET_x$ can be more or less accurately estimated by using simulation models, estimating the $ET_a$ with high precision is challenging because of the complexity of an agro-ecological environment. This makes Eq.~\ref{eq:yieldl} difficult to solve accurately in practice. 
Accurately estimating $ET_a$ usually involves field trials that are time-consuming, expensive, and not scalable over large areas~\cite{pereira2021standard}.   
Additionally, simulation models are often restricted to coarse spatial resolution due to computational complexity and low-resolution weather data, making in-field management (e.g., irrigation) difficult. \\

In this work, we address this limitation. We hypothesize that we can estimate the actual ET ($ET_a$) and the susceptibility to water scarcity ($K_y$) using a neural network (NN). 
This serves a dual purpose. First, estimating $ET_a$ and $K_y$ using a NN mitigates the limitations of the commonly used and inaccurate simulation models. 
Thus, we can approximate ET at a higher spatial resolution than the simulation models with high precision. Here, the multispectral satellite imagery and the ground truth yield data enable a form of super-resolution. Additionally, this data helps to capture the complex relationship between crop yield and water stress. 
As demonstrated in~\cite{allen2007satellite}, $ET_a$ can be approximated using satellite imagery and remote sensing technologies, which provides substance for our model.
Secondly, by using the yield response to water function (Eq.~\ref{eq:yieldl}), we enable yield prediction with physical consistency, thereby addressing the limitation of commonly employed black-box ML models. 

\section{Methodology}\label{sec:methodology}
Given the input data $x \in \mathcal{X}$, where $x$ is a multivariate time series with $T$ time steps, $x = (x^t)_{t=0}^T$, the actual target yield data $y_a \in \mathcal{Y}_a$, and the maximum evapotranspiration $ET_x \in \mathcal{ET}_x$, given as a time series with $T$ time steps, $ET_x = (ET_x^t)_{t=0}^T$. 
We aim to learn a function $f^\theta(x) = [ET_a, K_y]$ by optimizing over the model's parameters $\theta$, such that:
$K_y \left(1 - \frac{ET_a}{ET_x}\right) = \left(1 - \frac{y_a}{Y_x}\right)$.
Note that both $ET_a = (ET_a^t)_{t=0}^T$ and $K_y = (K_y^t)_{t=0}^T$ are estimated over time. The initial condition is defined by $y_l(0)=0$, and the final condition is defined by $y_l(T) = \left(1 - \frac{y_a}{Y_x}\right)$. The cumulative yield loss at the end of the time series is then given as:
\begin{equation}\label{eq:integral}
    y_l = \int_0^T K_y^t \left(1 - \frac{ET_a^t}{ET_x^t}\right) \, dt 
    \approx \sum_{t=0}^T  K_y^t \left(1 - \frac{ET_a^t}{ET_x^t}\right),
\end{equation}
providing us with accurate and physically consistent estimations of the relative yield loss. The final prediction of the actual yield is given by:
\begin{equation}\label{eq:yield}
    \hat{y}_a = y_l \cdot Y_x.
\end{equation}
Therefore, optimizing the model to approximate $\hat{y}_a $ improves its estimations of $ET_a$. We derive $Y_x$ in section~\ref{sec:data}. 
\subsection{Optimization}
We propose enforcing physical constraints, such that the reduction in ET corresponds to the observed reduction in yield. We assume that simulated $ET_x$ values are sufficiently accurate. 
This is supported by previous studies \cite{cai2007estimating}, allowing us to predict $ET_a$ and $K_y$ to achieve an accurate solution. For this, $\forall t \in [0,T]: 0 \leq ET_a^t \leq ET_x^t$ must hold. We describe the process of generating $ET_x$ values with a simulation model in section~\ref{sec:data}.
To estimate the yield reduction at time step $t$ as a function of a reduction in ET, a two-component loss term is proposed, consisting of a data-dependent part $\mathcal{L}_l$ and a physics part $\mathcal{L}_{phys}$:
\begin{align}
    \mathcal{L}_{\text{total}} &= \lambda_1 \mathcal{L}_{l} + \lambda_2 \mathcal{L}_{\text{phys}} \label{eq:loss} \\
    \mathcal{L}_{l} &= \mathbb{E} \left[ (\hat{y}_a - y_a)^2 \right] \label{eq:mse} \\
    \mathcal{L}_{\text{phys}} &= 
    \mathbb{E} \Big[ 
        \underbrace{1_{\{ET_a < 0\}} \cdot (ET_a)^2}_{\text{lower bound penalty}} + \nonumber \\
        &\phantom{= \mathbb{E} \Big[ }
        \underbrace{1_{\{ET_a > ET_x\}} \cdot (ET_a - ET_x)^2}_{\text{upper bound penalty}} + \nonumber \\
        &\phantom{= \mathbb{E} \Big[ }
        \underbrace{1_{\{0 \leq ET_a \leq ET_x\}} \cdot (ET_a - ET_x)^2}_{\text{within bounds MSE}} 
    \Big] \label{eq:phy}
\end{align}
Here, $1\{\cdot\}$ is an indicator function, which equals 1 if the condition inside the braces is true and 0 otherwise. The data-dependent component pushes the network to learn $ET_a$ such that by using Eq.~\ref{eq:yield} the predicted yield ($\hat{y}_a$) is close to the actual yield ($y_a$) of the ground truth data. The second component forces the network to maintain $ET_a$ values bounded between [0, $ET_x$] while consistently close to $ET_x$ to solve Eq. \ref{eq:yieldl}. Moreover, $\lambda_1$ and  $\lambda_2$ are hyperparameters that controls the weighting of both terms.
\subsection{Dataset}\label{sec:data}
\paragraph{Ground Truth Yield Data}
For training and evaluation, a publicly available yield data set is used~\footnote{https://www.research-collection.ethz.ch/handle/20.500.11850/581023}. This data set was presented by \citet{perich2023pixel} and we refer to it as \textit{SwissYield}. The data set comprises 54,098 yield samples from 54 yield maps of cereal crops recorded in Switzerland between 2017 and 2021. The data is characterized by georeferenced data points, collected by combine harvesters, containing information about the yield in tons/hectare [t/ha] in fine-scale resolution. After applying data preprocessing, \citet{perich2023pixel} rasterized the target data to $10 \times 10 \text{ m}$ pixel resolution using the geocube~\footnote{https://pypi.org/project/geocube/} package with linear interpolation. For more details, we refer the reader to \cite{perich2023pixel}.
For simplicity, we define the maximum yield sample as the maximum potential yield across the entire dataset: 
\begin{equation}
    Y_x = \max(\mathcal{Y}_a).
\end{equation}
\paragraph{Simulated Evapotranspiration}\label{sec:sim_data}
For each field, we generate simulation data for the crop $ET_x$ that is used for additional network regularization. 
For the simulations, we employ the FAO paper-56 \cite{allen1998crop} that simulates $ET_x$ over time. We use a publicly available Python implementation~\footnote{https://github.com/kthorp/pyfao56}~\cite{thorp2022pyfao56}. The relevant meteorological data that is described in Eq.~\ref{eq:ET_o} is acquired from the ERA5 global reanalysis program~\cite{hersbach2020era5} for every data sample. Features that are not available at 2 m height are adjusted following~\cite{allen1998crop}. Relevant soil data is collected from the SoilGrids~\cite{poggio2021soilgrids} and Hihydrosoil~\cite{simons2020hihydrosoil} project for every sample. Crop-specific parameters are taken from~\cite{allen1998crop,pereira2021standard}. For detailed information about the implementation, we refer to \cite{thorp2022pyfao56}.

\begin{figure}[!t]
    \centering
    \includegraphics[width=.95\linewidth]{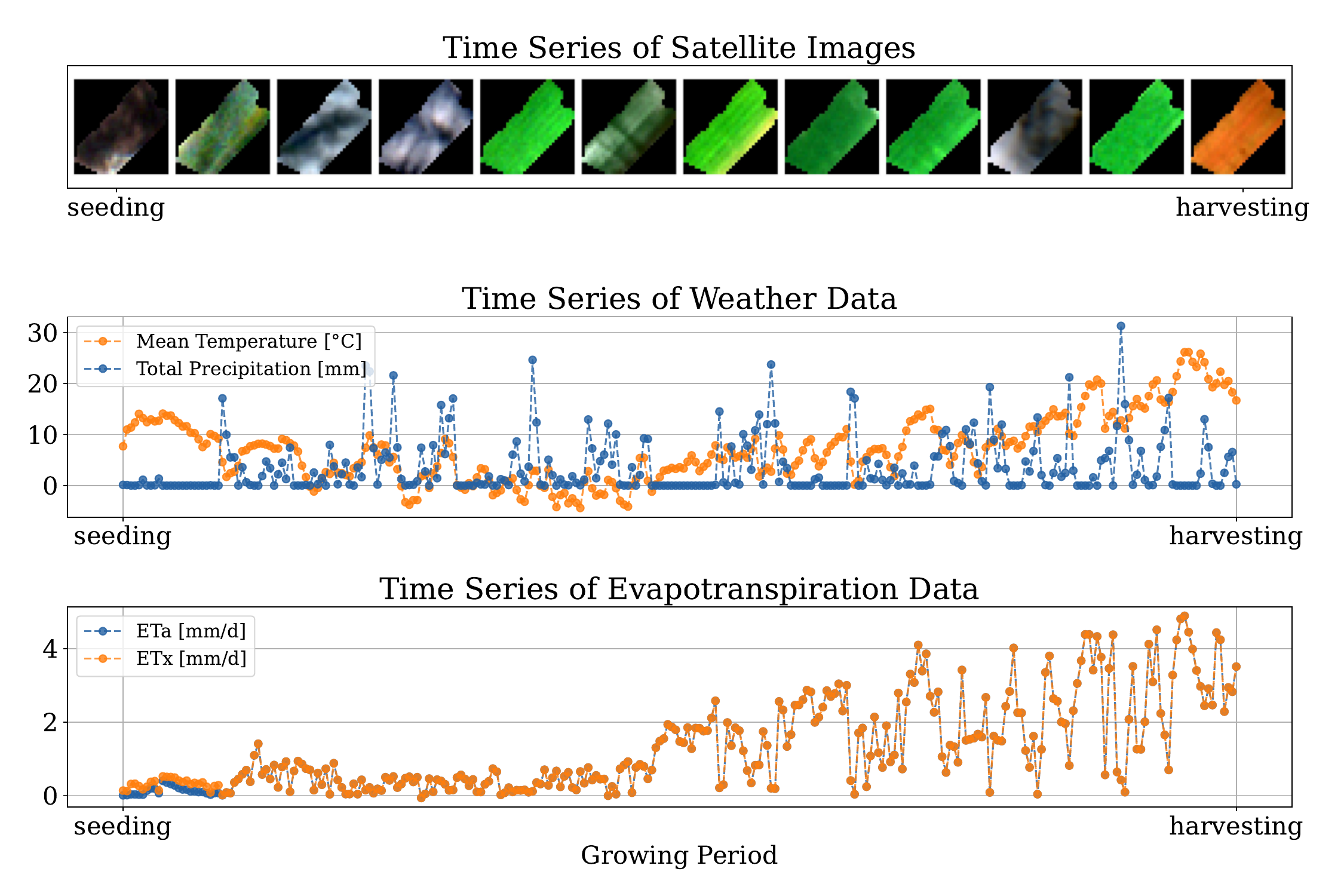}
    \caption{Example time series from seeding to harvesting of the training and simulation data of a randomly selected field. \texttt{Top:} Time series of satellite images in RGB. \texttt{Center:} Time Series of meteorological data. \texttt{Bottom:} Simulated maximum and actual evapotranspiration (ET). \textbf{Takeaway:} Maximum and actual ET values are almost consistently identical throughout the season in this example. }
    \label{fig:training_data}
\end{figure}

\paragraph{Training Data}
As the model input, time series data from the Sentinel-2 (S2) satellite mission is used. S2 data provides multispectral information in a wide range of the electromagnetic spectrum with a high revisit time of approximately 5 days at the Equator and a spatial resolution of up to $10 \times 10 \text{ m}$. S2 largely contributed to the recent success in EO by supporting services and applications of agriculture, land monitoring, climate change, and risk mapping. \\
Additionally, meteorological data is acquired per field and incorporated into the network to better account for extreme environmental conditions. More specifically, the total precipitation and the minimum and maximum temperature are used, derived from ERA5 global reanalysis~\cite{hersbach2020era5}. Since meteorological and S2 data have different temporal and spatial resolutions, data preprocessing and data fusion are required. 
Data modalities are fused at the input level using the raw time series of S2 images by aggregating weather features between S2 time steps, following~\cite{pathak}. The characteristics of the final data set are given in Table~\ref{tab:data}.
Figure \ref{fig:training_data} illustrates a time series of input data and simulation data for a randomly selected field. Note that both the meteorological and simulation data are characterized by daily measurements but lack spatial information. Moreover, S2 imagery has fewer time steps available compared to meteorological and simulation data. Additionally, we highlight that ET correlates with the temperature and the vegetation that is depicted in the S2 images. More importantly, we emphasize that only minor differences are observed between the maximum and actual ET, highlighting the challenge of accurately solving the yield response to water function (Eq.~\ref{eq:yieldl}) at fine-scale resolution using simulation models alone.
\begin{table*}[!t]
    \centering
    \caption{Overview of the yield dataset and its characteristics. }
    \resizebox{.95\linewidth}{!}{%
\begin{tabular}{l|cccccccccc}
\hline
\textbf{Dataset}&
  \textbf{Countries} &  \textbf{Crops} &  \textbf{Years} &  \textbf{Fields}  & \textbf{Pixel-Level} & \textbf{Samples} & \textbf{Time Steps} & \multicolumn{2}{c}{\textbf{Resolution}} & \textbf{Features}  \\ 
  \cmidrule(lr){9-10} 
                  &                  &         &   &  &  &  &  & Spatial   & Temporal & \\ \hline

SwissYield \cite{perich2023pixel}  & 1          & 1          & 2017–2021           &   54                  & \checkmark & 54,098  & 16-55   &     10m       & ~5 days          & 14                      \\ \hline
\end{tabular} 
    }
    \label{tab:data}
\end{table*}

\subsection{Architecture}
For the implementation, we leverage a recurrent neural network (RNN) architecture, more specifically a long short-term memory (LSTM)~\cite{LSTM} backbone with 2 layers, where each hidden state is passed to a sequential layer with 128 hidden units, incorporating a linear layer, batch normalization, and dropout of 0.2. Finally, two linear layers are incorporated with a single output channel each, predicting $K_y^t$ and $ET_a^t$, respectively. 
We refer to this network as physics-guided LSTM (PG-LSTM). 
LSTM networks have shown great performance in sequence forecasting even over Transformer models~\cite{Sonata_lstm_transformer}, also in crop yield prediction~\cite{helber2023crop}. However, since LSTM models can struggle with very long and multidimensional sequences, we evaluate the inclusion of an attention mechanism, as proposed in \cite{he2023physics}. More specifically, we employ the scaled dot-product attention~\cite{atten_transformer_vasani}, as it has been used in crop yield prediction~\cite{pathak}. We refer to this model as PG-LSTM$^{attn}$. A schematic overview of the training and inference scheme of the proposed method is given in Figure~\ref{fig:model_training_inference}. The data is processed pixel-wise and fused at the input level. Moreover, simulated $ET_x$ values are used in the optimization loss to guide the training. At each time step, the model produces an estimation of $ET_a$ and $K_y$ which is then used to calculate the yield through the yield response to water function. S2 images and ground truth yield data enable the estimation of $ET_a$ and crop yield at the pixel resolution.  \\
\textbf{Uncertainty:} Since we aim to estimate two components ($ET_a$ and $K_y$) this problem becomes ill-defined, increasing the solution space and potentially introducing uncertainty in the predictions. Therefore, we must account for the uncertainty in the model. We do so by using a deep ensembles approach~\cite{lakshminarayanan2017simple}. Deep ensembles are a powerful tool for uncertainty estimation that affectively approximate Bayesian marginalization while also improving accuracy and out-of-distribution robustness. Deep ensembles are becoming the gold standard in estimating well-calibrated predictive distributions~\cite{wilson2020bayesian}. We train 10 separate ensemble members to estimate the uncertainty in the proposed method.  \\
 
\subsection{Experimental Setup}
For each experiment, a K-fold cross-validation (K=10) is performed, where we present the result as the average across folds. Moreover, we perform a Leaf-One-Year-Out (LOYO) cross-validation scenario, where one year is held out during training and used only for evaluation to estimate the temporal transferability.  
To evaluate the yield prediction performance, standard regression metrics are used. This includes the coefficient of determination ($R^2$-score), mean absolute error (MAE), mean absolute percentage error (MAPE), root-mean-square error (RMSE), and the Bias. For qualitative evaluation, we visually evaluate the predicted in-field variability, low spatial prediction error, and a match between predicted and target distributions. Additionally, we visually evaluate the predicted actual ET using agricultural experts. \\
We compare the PG approach against several state-of-the-art models for crop yield prediction without any physical components and regularization. This includes a LSTM and a Transformer architecture. Moreover, we include a simple linear regression model. For the LSTM model, we implement the architecture as described in \cite{pathak}. The Transformer model is based on the Transformer Encoder as introduced by~\cite{vaswani2017attention} with two layers and a convolution operation that maps the input dimension to the hidden size of 128. Finally, a multilayer perceptron with ReLU activation predicts the final yield. Between these operations, Layer Normalization is applied. \\

For all NN models, training is conducted using the ADAM optimizer for a maximum of 100 epochs. The learning rate is set to 0.001, and the batch size is 512. A reduce-on-plateau learning rate scheduler is employed during training. For regularization, early stopping is applied if there is no improvement on the validation set for 10 consecutive epochs. 
\subsection{Results}
\subsubsection{Yield Prediction}
\begin{figure}[!t]
    \centering
    \includegraphics[width=.99\linewidth]{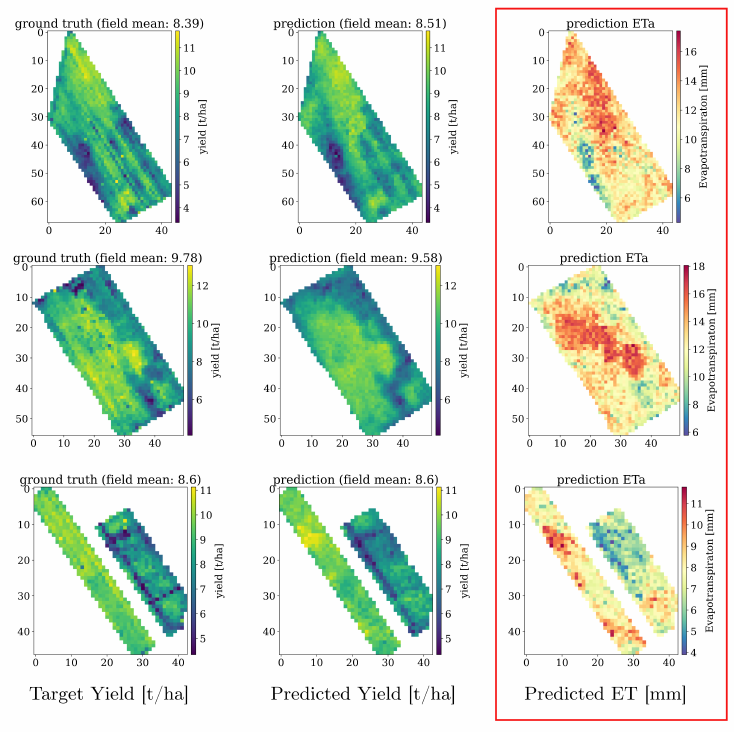}
    \caption{Example predictions of an entire field. Each row represents a single field. \texttt{Left:} Ground truth yield map, \texttt{center:} predicted yield map, \texttt{right:} predicted actual ET. \textbf{Highlight:} The actual ET originally had no spatial resolution at the pixel level. Now we observe a clear correlation between the actual ET and the predicted yield, such that high ET values correlate with high yield values.}
    \label{fig:predicted_field}
\end{figure}

We first investigate whether the proposed model is competitive in crop yield prediction by evaluating the yield prediction performance against state-of-the-art models. The quantitative results are presented in Table~\ref{tab:model_comparision}. For our method, we use the last time steps for evaluation and report the average across all ensemble members. Interestingly, both PG models outperform all state-of-the-art models, including Transformer, LSTM, and Linear Regression by a great margin. For instance, the PG-LSTM$^{attn}$ model improves 9 percentage points (p.p.) over the Transformer model in $R^2$-score and 2 p.p. over the standard LSTM model. As expected, the Linear Regression model exhibits the poorest performance but still achieves a noteworthy $R^2$-score of 0.7. Comparing both PG-LSTM$^{attn}$ and PG-LSTM, we observe only minor differences, however, with a slightly better performance of the PG-LSTM$^{attn}$. \\
A qualitative example field is depicted in Figure~\ref{fig:predicted_field} for the PG-LSTM$^{attn}$. 
Notably, the model exhibits high in-field variability that closely aligns with the target data.
\begin{table}[!t]
    \centering
    \caption{Overview table of yield prediction performance of different models. \textbf{Highlight:} The physics-guided models outperform state-of-the-art methods on main metrics. }
    \resizebox{.95\columnwidth}{!}{%
    \begin{tabular}{l|ccccc}
\hline
\multicolumn{1}{c|}{Option} & 
\begin{tabular}[c]{@{}c@{}}$R^2$-score\\ -\end{tabular} &
\begin{tabular}[c]{@{}c@{}}MAE\\ t/ha\end{tabular} &
\begin{tabular}[c]{@{}c@{}}MAPE\\ \% \end{tabular} &
\begin{tabular}[c]{@{}c@{}}RMSE\\ t/ha\end{tabular} &
\begin{tabular}[c]{@{}c@{}}BIAS\\ t/ha\end{tabular} 
\\ \hline
PG-LSTM$^{attn}$ & \textbf{0.82} & \textbf{0.59} & \textbf{0.11} & \textbf{0.86} &   -0.01                  \\
PG-LSTM           & 0.81 & 0.59 & \textbf{0.11} & 0.87  & 0.07              \\
Transformer       & 0.73 & 0.74 & 0.14 & 1.05  & 0.41                \\
LSTM              & 0.8  & 0.62 & 0.12 & 0.9   & 0.04                 \\
Linear Regression & 0.7  & 0.81 & 0.17 & 1.1  & \textbf{0} \\ \hline
\end{tabular}
 
    }
    \label{tab:model_comparision}
\end{table}
\subsubsection{Physical Consistency} 
Furthermore, we highlight in Figure~\ref{fig:predicted_field} that the predicted $ET_a$ values now have a spatial resolution of $10 \times 10 \text{ m}$, compared to a single value before. We evaluate this for the entire time series.  
We notice that the predicted $ET_a$ values correlate significantly with the target yield data at later time steps, suggesting that the model learned the relationship between yield and ET at the pixel-level. More specifically, high $ET_a$ values correlate with a higher yield as desired. Conversely, areas with lower $ET_a$ correspond to lower yields, indicating crop stress. This is highlighted by the significant correlation of the final prediction with the ground truth yield data. This underlines the capabilities of estimating the crop yield reduction over time through learning the important features of the crop water use.
\begin{figure*}[!t]
    \centering
    \includegraphics[width=.99\linewidth]{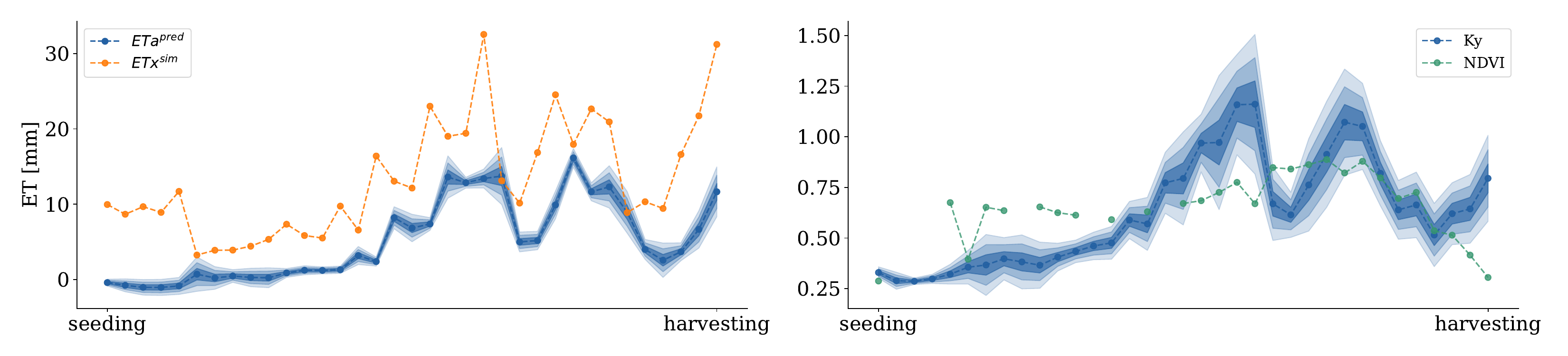}
    \caption{Single field example of the predicted agronomic properties. \texttt{Left:} Temporal visualization of the simulated maximum $ET_x$ and learned actual $ET_a$. \texttt{Right:} Temporal visualization of the estimated susceptibility to water scarcity ($K_y$) and the NDVI. Predictions are illustrated with $\pm2\sigma$ to account for uncertainty. \textbf{Highlight:} Learned actual $ET_a$ and $K_y$ values follow key physical principles.}
    \label{fig:ET_single_field}
\end{figure*}
\begin{figure*}[!t]
    \centering
    \includegraphics[width=.99\linewidth]{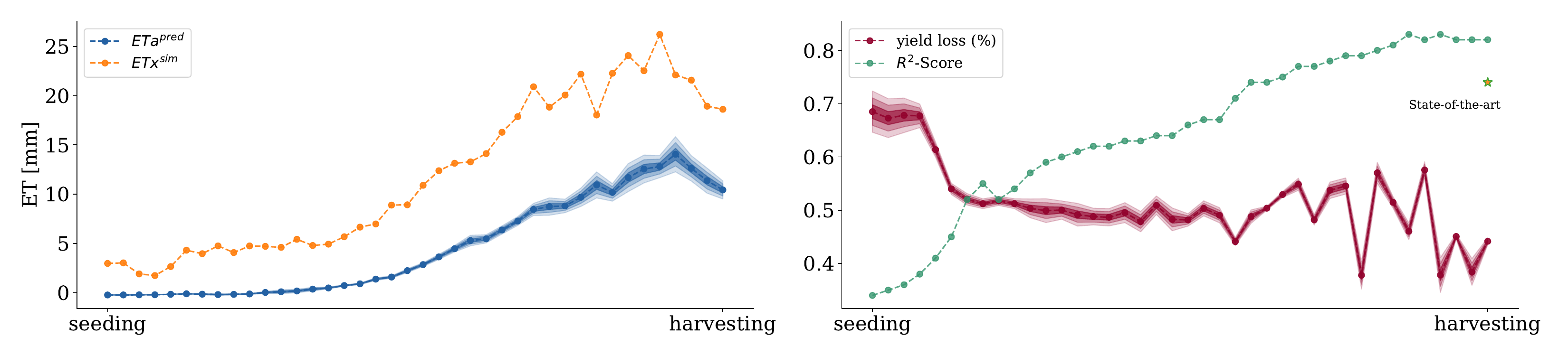}
    \caption{ Visualization of agronomic properties of the entire dataset. 
    \texttt{Left:} Temporal illustration of the simulated \underline{maximum} ($ET_x^{sim}$) and learned \underline{actual} ($ET_a^{pred}$) crop evapotranspiration (ET), an indicator of drought stress. 
    \texttt{Right:} The reduction in $ET_a$ compared to $ET_x$ is used to derive the reduction in yield (yield loss) using the yield response to water function~\cite{doorenbos1979yield}. The model accuracy is expressed by the $R^2$-score over time. Predictions are illustrated with $\pm2\sigma$ using a deep ensemble approach. \textbf{Highlight:} The estimated yield loss correlates with the learned drought stress while outperforming state-of-the-art models. }
    \label{fig:prediction_overall}
\end{figure*}
In the following, we evaluate the physical consistency of the proposed approaches. 
In Figure~\ref{fig:ET_single_field} we illustrate the sequential estimations of the simulated $ET_x$ and predicted $ET_a$ values for the same field as depicted in Figure~\ref{fig:predicted_field}. On the right, we show the estimated $K_y$ values alongside the normalized difference vegetation index (NDVI). The NDVI is derived from the satellite imagery and serves as an indicator of vegetation density and plant health. It can be used to evaluate the physical consistency of the models, as an increase in ET should be accompanied by high NDVI values until senescence. We display the mean prediction for $ET_a$ and $K_y$ of all ensemble members over all field samples, together with a buffer of $\pm 2\sigma$ to assess the temporal uncertainty. First, we notice that the uncertainty for the $ET_a$ values is lower compared to the $K_y$ values. This can be explained by the fact that $K_y$ acts as a free parameter, which can result in higher uncertainties. 
We observe strong consistency between the simulated $ET_x$ and predicted $ET_a$ values, following the conditions defined in Sec.~\ref{sec:methodology}, indicating that the model captured important agronomic properties. $ET_a$ values are consistently lower than $ET_x$, indicating yield-limiting conditions, such as water scarcity. Nevertheless, a consistent increase in ET over the growing period is observed, which correlates with an increase in NDVI and $K_y$, indicating greater susceptibility to water scarcity at later growth stages, as reported in~\cite{steduto2012crop}. \\ 
In Figure~\ref{fig:prediction_overall} we display the simulated $ET_x$ and predicted $ET_a$ values averaged over the dataset. Additionally, the model performance over time, expressed in $R^2$-score, is shown alongside the estimated yield loss. 
We observe similar behavior as in the single field example. A consistent increase in ET over the growing period is observed, with a reduction in predicted yield loss and an improvement in yield prediction performance. We highlight that the predicted yield loss negatively correlates with the predicted ET, with a Pearson correlation of -0.45. This indicates that the model learned the relationship between reduction in ET and the corresponding reduction in yield. 

\subsubsection{Ablation Studies}
\begin{table}[!t]
    \centering
    \caption{Overview of the model performance without estimating the crop susceptibility to water scarcity ($K_y$). \textbf{Takeway:} Using constant $K_y$ values from previous research, still produces accurate yield estimates.}
    \resizebox{.95\columnwidth}{!}{%
    \begin{tabular}{l|ccccc}
\hline
\multicolumn{1}{c|}{Option} & 
\begin{tabular}[c]{@{}c@{}}$R^2$-score\\ -\end{tabular} &
\begin{tabular}[c]{@{}c@{}}MAE\\ t/ha\end{tabular} &
\begin{tabular}[c]{@{}c@{}}MAPE\\ \% \end{tabular} &
\begin{tabular}[c]{@{}c@{}}RMSE\\ t/ha\end{tabular} &
\begin{tabular}[c]{@{}c@{}}BIAS\\ t/ha\end{tabular} 
\\ \hline
PG-LSTM$^{attn}$    & 0.26          & 1.28         & 0.21          & 1.73              & -0.18                \\
PG-LSTM             &  0.74         &  0.72         & 0.14         &   1.02            & 0.13               \\ \hline
\end{tabular}
 
    }
    \label{tab:NoKy}
\end{table}
\paragraph{Assessing the importance of the yield response factor:}
\begin{table}[!t]
    \centering
    \caption{Performance overview for the Leaf-One-Year-Out cross-validation scenario. \textbf{Takeaway:} All models struggle to predict unknown years.}
    \resizebox{.95\columnwidth}{!}{%
\begin{tabular}{lccccc|ccccc}
\hline
\multicolumn{1}{c}{\multirow{2}{*}{Option}} &
  \multicolumn{5}{c|}{\begin{tabular}[c]{@{}c@{}}$R^2$-score\\ -\end{tabular}} &
  \multicolumn{5}{c}{ \begin{tabular}[c]{@{}c@{}}RMSE\\ t/ha \end{tabular}} \\ \cline{2-11} 

\multicolumn{1}{c}{} &
  \multicolumn{1}{l}{2017} &
  \multicolumn{1}{l}{2018} &
  \multicolumn{1}{l}{2019} &
  \multicolumn{1}{l}{2020} &
  \multicolumn{1}{l|}{2021} &
  \multicolumn{1}{l}{2017} &
  \multicolumn{1}{l}{2018} &
  \multicolumn{1}{l}{2019} &
  \multicolumn{1}{l}{2020} &
  \multicolumn{1}{l}{2021} \\ \hline
PG-LSTM$^{attn}$  &
  0.09 &
  -0.38 &
  0.19 &
  \textbf{0.1} &
  \textbf{0.29} &
  1.87 &
  1.93 &
  1.6 &
  2.24 &
  2.92 \\
PG-LSTM  &
  0.14 &
  0.04 &
  \textbf{0.41} &
  -0.71 &
  0.26 &
  1.82 &
  1.48 &
  \textbf{1.38} &
  2.16 &
  \textbf{1.89} \\
Transformer &
  0.31 &
  \textbf{0.3} &
  0.31 &
  -0.31 &
  -0.61 &
  1.63 &
  1.27 &
  1.48 &
  1.89 &
  2.79 \\
LSTM &
  0.24 &
  0.29 &
  0.25 &
  -0.68 &
  -0.91 &
  1.71 &
  \textbf{1.27} &
  1.55 &
  \textbf{2.14} &
  3.03 \\
Linear Regression &
  \textbf{0.37} &
  -0.36 &
  -0.3 &
  -1.63 &
  -0.99 &
  \textbf{1.56} &
  1.76 &
  2.04 &
  2.68 &
  3.1 \\ \hline
\end{tabular}
 
    }
    \label{tab:loyo}
\end{table}
To assess the importance of the free parameter $K_y$, we illustrate in Table~\ref{tab:NoKy} the model performance of the PG methods that estimate only the $ET_a$ values. For $K_y$ a constant of 1.05 is defined as provided in~\cite{steduto2012crop}. Unexpectedly, the PG-LSTM significantly outperforms the PG-LSTM$^{attn}$ by 48 p.p. in $R^2$ and by 0.71 t/ha in RMSE. However, PG-LSTM still achieves competitive yield prediction performance with an $R^2$ of 0.74, thereby outperforming the Transformer and Linear Regression model, as evidenced before in Table~\ref{tab:model_comparision}. Moreover, this suggests that while the $K_y$ is a meaningful parameter to estimate, previously reported values are still providing sufficient information to study the relationship between ET and yield reduction. 
\paragraph{Temporal Transferability:}
Table~\ref{tab:loyo} presents the temporal transferability of all methods under the LOYO cross-validation scenario. Notably, all methods exhibit significantly lower performance when applied to unseen years, with a maximum $R^2$ of 0.41 of the PG-LSTM in 2019. In particular, the Linear Regression model demonstrates the poorest temporal transferability. In contrast, PG-LSTM and PG-LSTM\textsuperscript{attn} achieve competitive performance compared to existing state-of-the-art methods, outperforming them in 2019 and 2021.

\section{Discussion}
This study investigated the inclusion of environmental water stress conditions into a ML to enhance crop yield prediction by ensuring physical consistency. While earlier studies have explored the integration of water stress into the data space, they have not explicitly enforced physical consistency through model regularization. In contrast, our work explicitly formulates crop yield as a function of water scarcity in the loss term. 
We find that this approach outperforms existing ML models for crop yield prediction \cite{pathak}. 
Compared to previous work, we demonstrated high explainability and that the model captured the complex relationship between crop stress and productivity.
Interestingly, we experimentally demonstrated that the estimation of crop water stress can be approximated at fine-scale resolution using multispectral satellite imagery and pixel-level yield data. The results indicate promising potential by overcoming limitations in both simulation models and data-driven yield prediction methods. \\
Although this approach represents a step towards more explainable and transparent yield prediction, limitations and simplifications remain that must be considered. Most importantly, the size of the datasets impedes the development and evaluation of more powerful and scalable models. 
More data is required to deeply assess the importance of this work, particularly data from water scarce regions, including more crop types, regions, and years. Limited data can cause dramatic performance degradation, such as demonstrated in the LOYO cross-validation scenario (Table \ref{tab:loyo}). This, moreover, underscores the need for further research and the integration of additional transfer learning techniques~\cite{helber2023crop}. Additionally, modeling ET is challenging~\cite{pereira2021standard}. Therefore, more attention must be devoted to calibrating the employed methods to avoid miscalibration and consequently, model overfitting.   
Ground truth samples of the actual ET are required to estimate the accuracy of this method. Future research should include more accurate ET values derived from field experiments and satellite data, such as evidenced in~\cite{allen2007satellite} to further reduce the uncertainty in our method. However, this work aims for high reproducibility, and publicly available datasets are scarce. 
\section{Conclusion}
Informed Neural Networks hold significant potential for crop yield modeling, offering enhanced adaptability to challenging environmental conditions. We presented a novel approach to modeling crop productivity under environmental constraints and demonstrated promising experimental outcomes. The presented approach supports industry, policymakers, and farmers in achieving more sustainable and resilient agriculture. 







\bibliography{ref}

\end{document}